# Is Contrasting All You Need? Contrastive Learning for the Detection and Attribution of AI-generated Text

**Lucio La Cava**[a,*], **Davide Costa**[a,1] and **Andrea Tagarelli**[a,**]

[a]DIMES Dept., University of Calabria, 87036 Rende (CS), Italy

**Abstract.** The significant progress in the development of Large Language Models has contributed to blurring the distinction between human and AI-generated text. The increasing pervasiveness of AI-generated text and the difficulty in detecting it poses new challenges for our society. In this paper, we tackle the problem of detecting and attributing AI-generated text by proposing WhosAI, a triplet-network contrastive learning framework designed to predict whether a given input text has been generated by humans or AI and to unveil the authorship of the text. Unlike most existing approaches, our proposed framework is conceived to learn semantic similarity representations from multiple generators at once, thus equally handling both detection and attribution tasks. Furthermore, WhosAI is model-agnostic and scalable to the release of new AI text-generation models by incorporating their generated instances into the embedding space learned by our framework. Experimental results on the TuringBench benchmark of 200K news articles show that our proposed framework achieves outstanding results in both the Turing Test and Authorship Attribution tasks, outperforming all the methods listed in the TuringBench benchmark leaderboards.

## 1 Introduction

In recent years, advancements in artificial intelligence (AI) have revolutionized various domains, including natural language processing (NLP), leading to the emergence of sophisticated text generation models. These AI-powered systems are capable of generating human-like text, ranging from simple sentences to complex narratives, with remarkable fluency and coherence [20]. Such advancements have attracted attention from the research community as well as industry and society at large, offering opportunities for enhancing communication, creativity, and productivity [7].

However, a pressing challenge comes alongside these advancements: distinguishing between AI-generated text and human-written text. As AI text generation models continue to improve in sophistication and realism, the ability to differentiate between AI-generated and human-generated content becomes increasingly crucial [37]. The implications of failing to discern between these two sources of text are profound and multifaceted, spanning various aspects of society, such as the preservation of truth, authenticity, and trustworthiness in online communication. With the proliferation of AI-generated content on social media, news platforms, and other digital channels, there is a growing risk of misinformation, manipulation, and deception [51, 9]. Without effective means of distinguishing between AI-generated and human-generated text, users may unwittingly consume and propagate false or misleading information, undermining the integrity of public discourse and decision-making processes.

Furthermore, the rise of AI text generation brings ethical and societal concerns about authorship, intellectual property rights, and accountability. As AI systems become increasingly proficient at mimicking human language and creativity [8], questions arise regarding the ownership and attribution of AI-generated content. Without clear guidelines and mechanisms for identifying the origin of text, issues might also arise about plagiarism, copyright infringement, and legal responsibility, posing challenges to established norms in intellectual property law and digital content creation.[2]

**Related work.** The remarkable boost in human-like text generation performances achieved by Large Language Models (LLMs) in recent years has determined a rising challenge in detecting whether and to what extent texts have been generated by humans or machines [21, 49, 39]. In this context, the "watermarking" paradigm rapidly gained attention [25, 53, 28, 50], as it allows embedding specific signals into generated texts that remain invisible to humans but are algorithmically detectable. Statistical learning methods also offer advanced solutions for detecting the authorship of texts. These include probabilistic models [31, 1, 47, 14], log rank information [38], perplexity [44], discourse motifs [24], and other statistical approaches [16, 40, 45].

More recently, we have witnessed the emergence of deep learning to detect or attribute AI-generated content, which stands as a promising body of research. Researchers have been exploiting LLMs to detect generated text [18, 46], using ChatGPT itself as a detector [2], or combining LLMs with topological aspects [43].

A very recent trend involves leveraging contrastive learning to handle textual information. Indeed, despite its origins in the computer vision domain, contrastive representation learning has been proven particularly effective in NLP contexts to improve research on semantic similarity related problems, such as text classification [33, 10], spotting hate-speech [23], unveiling intents [55], and eventually detecting AI-generated text through domain adaptation [3] or domain adversarial training [4].

Despite the advancements in research on detection of AI text generation, each of the above mentioned approaches faces significant challenges. Watermarking approaches are conditioned by the con-crete possibility of watermarking a given text, leaving the detection of non-watermarked texts an open issue. Statistical learning methods typically require access to the models' internals or to information that might be unavailable, limiting their applicability. Yet, more importantly for the sake of comparison with our proposed approach, existing contrastive-learning-based methods require or are better developed when learning a separate model for each AI generator.

---

[*] lucio.lacava@dimes.unical.it
[**] Corresponding Author, andrea.tagarelli@unical.it
[1] Davide Costa was affiliated with the DIMES Dept. of the University of Calabria until the time of submission of this paper.

[2] *Can one really spot if the text in the Introduction was written by a human or an AI text-generation model?*

**Contributions.** In light of the above remarks, we aim to fill a gap in detecting and attributing AI-generated text by proposing WhosAI, a novel learning framework that leverages deeply contextualized dense representations of textual data as core of a *contrastive triplet learning* architecture to address binary/multi-class prediction tasks of authorship attribution of texts written by humans or AI text-generation models. The key idea underlying WhosAI is to integrate the power of Transformer-based pretrained language models (PLMs) into a framework of similarity learning optimizing a contrastive triplet loss function to learn deep semantic subspaces that maximize the cohesiveness of groups of similar texts and the separation of groups of dissimilar texts. Compared to existing solutions for detecting AI-generated text, WhosAI features the following key advantages:

- WhosAI does not require editing texts (unlike in watermarking methods), or accessing AI generation models' internals, and does not make any assumption on linguistic features that might be exhibited by particular text generators, or on any degree of open-endedness in text generation, thus WhosAI can *deal with on any type of texts*;
- WhosAI is conceived to be *versatile* w.r.t. the particular PLM used at the core of the learning framework, and is *general-purpose*, as it does not require training separate models for different tasks or even generators, overcoming a major issue of existing approaches based on contrastive learning;
- The contrastive learning approach in WhosAI makes is *model-agnostic* and *scalable* to the release of new AI text-generators; indeed, is it sufficient to add new data to the training set to enable the proposed framework to generalize to new text generators.

The significance of WhosAI has been demonstrated based on a thorough evaluation on the widely recognized *TuringBench* benchmark dataset, comprising 200K articles that are either human-written or generated by 19 different AI text-generation models. In this challenging context, WhosAI achieves excellent results in terms of both classification performance and internal validity criteria, outperforming all the methods appearing in the benchmark's leaderboard, for both the *Turing Test* and *Authorship Attribution* tasks.

## 2 Problem Statement

We are given a set of discrete labels (categories) $\mathcal{C} = \{c_j\}_{j=1}^M$, with $M \geq 2$, and a collection of text data objects $\mathcal{D} = \{D_i\}_{i=1}^N$, such that each text object in $\mathcal{D}$ is assigned to one of the categories in $\mathcal{C}$. The semantics of such categories refer to information on the originator of a written text, which is assumed to be either a human or a machine, i.e., an AI model for text generation; hence, in this setting, the authorships of the texts in $\mathcal{D}$ are a priori known.

The problem we are interested in is generally to learn a model, supervisedly trained on $\langle \mathcal{D}, \mathcal{C} \rangle$, that can predict the category from $\mathcal{C}$ for any given text data whose authorship is unknown. Specifically, we address two supervised learning problems: ($i$) A binary classification task, known as **Turing Test** (TT), which requires to predict whether the author of a text is a human or an AI text-generator, and ($ii$) A multi-class classification task, known as **Authorship Attribution** (AA), which requires to predict exactly who is the author of a text, choosing between a human or an AI text-generator.

In line with the related literature, our setting does not differentiate between human authors in either task (i.e., 'human' always corresponds to one class in $\mathcal{C}$), whereas the identity of a particular AI text-generator must be unveiled for the Authorship Attribution task only, therefore $M-1$ categories are available that correspond to either *any* AI text-generator (for Turing Test) or *a specific* AI text-generator (for Authorship Attribution).

It is worth emphasizing that these tasks, particularly the Authorship Attribution, pose in principle two challenges:

- First, the available AI text-generator categories might not necessarily be regarded as classes of document representations that unlikely share their linguistic feature subspaces; roughly speaking, two texts generated by different AI models, or even the same model with different parametrization, could be hardly distinguishable form each other. This particularly holds in our setting since the AI text generation is assumed to be *open-ended*.
- Second, and more importantly, it is supposed that the number of AI text-generators will keep growing, however a classification model trained to recognize the authorship of a text would have to be retrained every time a new AI text-generator is added to the document database.

The above challenges can be faced if the problem under study is switched to a *similarity learning* problem as a core component for the ultimate goal of binary/multi-class prediction.

## 3 Background

**Transformer-based Pre-trained Language Models** (PLMs) are the well-established NLP tools to build deeply contextualized text-representation learning models. Given a text data $D_i \in \mathcal{D}$, a token sequence $T_i = [\tau_{i,1}, \ldots, \tau_{i,|T_i|}]$ is produced as initial representation of $D_i$ through a *tokenization* process typically associated with a PLM. Each token sequence is deeply contextualized by mapping it onto a dense, relatively low dimensional space of size $f$, based on the PLM. The resulting output is the *token embeddings* of $D_i$, denoted as $\mathsf{PLM}(T_i) \in \mathbb{R}^{f \times |T_i|}$. Eventually, a pooling function $pooling(\cdot)$ is applied to the token embeddings of each object $D_i$ to yield a single embedding vector $\mathbf{h}_i$ of size $f$:

$$\mathbf{h}_i = pooling(\mathsf{PLM}(T_i)) \in \mathbb{R}^f. \quad (1)$$

Typically, this pooled output is an average embedding over all token embeddings of a data object. The embeddings $\mathbf{h}_i$ are commonly referred to as *sentence embeddings*.

**Similarity Learning.** The deeply contextualized representations produced by a PLM lend themselves particularly suited to enable semantic comparisons between the input text objects. In this respect, we want to explicitly model and leverage the similarity space induced from the sentence embeddings. *Similarity learning* aims to train a model to distinguish between similar and dissimilar pairs of objects. More specifically, if we consider objects whose relative similarity follows a predefined order – i.e., for any triplet of objects, the first object is assumed to be more similar to the second object than to the third object – the goal becomes to learn a *contrastive loss* function, so that it favors small distances between pairs of objects labeled as similar, and large distances for pairs labeled as dissimilar. This is certainly our case since it is expected that a human-written text to be similar to another human-written text than an AI-generated text, or texts generated by the same AI model to be similar to each other than to texts generated from other AI models.

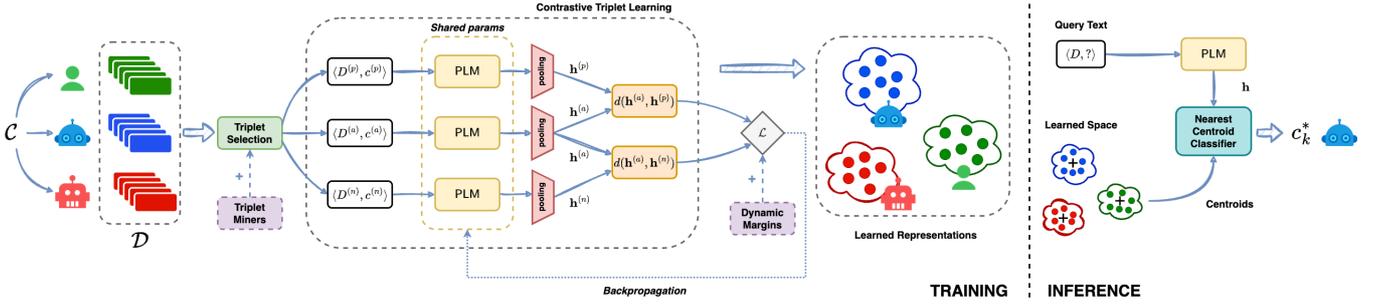

**Figure 1.** Overview of our proposed WhosAI learning framework, at training time (left) and inference time (right).

Contrastive learning is often performed by using a *Siamese Network* architecture [5], which contains two PLM instances sharing the same weights while being trained in parallel on two input objects to compute comparable outputs. When using a contrastive triplet loss, Siamese Network is commonly referred to as *Triplet Network*.

## 4 The WhosAI Framework

**Overview.** We propose WhosAI, a deep learning framework for the detection and attribution of open-ended texts generated by AI models vs. human-written texts. Figure 1 shows a schematic illustration of the main components and data flows of WhosAI.

WhosAI is conceived to be trained on text data with associated labels expressing authorship as either human or an AI text-generation model. The framework is comprised of three key elements: (i) a PLM, which is charge of learning deeply contextualized representations (embeddings) of the text data, in an unsupervised fashion, (ii) a Triplet Network architecture, which is designed to perform contrastive learning to induce a similarity space of the PLM embeddings, and (iii) a nearest centroid classification model, which is in charge of predicting the authorship category for any query text.

During the training phase, WhosAI builds a deep semantic representation space whereby different regions correspond to features of human-written texts as well as distinct AI text-generators. The contrastive learning strategy allows for capturing the underlying similarity structure and relations within the data objects, such that the deeply contextualized embeddings produced by a PLM encoder will be grouped together when they correspond to the same author and will be kept separated when they correspond to different authors. Moreover, as a byproduct, the similarity learned space facilitates the learning of the decision boundary for our classification objective of determining the class of previously unseen texts; in this setting, our choice of a nearest centroid classifier turns out to be a highly efficient yet effective way to perform authorship prediction.

WhosAI is designed to be *versatile* and *modular*. Versatility mainly refers to the possibility of choosing alternative PLMs as core component of the Triplet Network, variants of the Triplet Network architecture, and alternative (instance-based) classification models. Moreover, WhosAI is modular in that enhanced methods are considered to improve specific aspects of the framework. In particular, these enhancements include (i) improving the efficiency and generalization capabilities of the contrastive learning component, (ii) refining the separation between classes corresponding to different text-creators in the learned space, and (iii) enhancing the robustness of the framework by corrupting the input textual data.

**Training.** Our training process starts with mining *triplets* $\langle D^{(a)}, D^{(p)}, D^{(n)} \rangle$ of text data objects from $\mathcal{D}$ to be fed into our triplet network. Such triplets are formed in such a way that, for a given *anchor* $D^{(a)}$, $D^{(p)}$ and $D^{(n)}$ are selected as *positive* and *negative* sample, respectively, i.e., such that $c^{(a)} = c^{(p)}$ and $c^{(a)} \neq c^{(n)}$, where symbols $c^{(\cdot)}$ are here used to denote the category associated with an anchor, positive or negative object.

The embeddings $\mathbf{h}^{(a)}, \mathbf{h}^{(p)}, \mathbf{h}^{(n)}$ of the anchor, positive and negative objects, respectively, are next computed according to Eq. 1. It should be noted that the text annotations, i.e., their associated categories, are not required when computing the embeddings, since the PLM is an unsupervised learner.

Given a triplet, the Triplet Network computes the distance between the embedding of the anchor object and the embedding of the positive object (*positive pair*), and the distance between the embedding of the anchor object and the embedding of the negative object (*negative pair*). The *triplet loss* minimizes the distance between an anchor and a positive, both having the same category, and maximizes the distance between the anchor and a negative of a different category:

$$\mathcal{L} = \sum_{\langle D^{(a)}, D^{(p)}, D^{(n)} \rangle} \max(d(\mathbf{h}^{(a)}, \mathbf{h}^{(p)}) - d(\mathbf{h}^{(a)}, \mathbf{h}^{(n)}) + \lambda, 0) \quad (2)$$

where $d(\cdot, \cdot)$ is a distance function and $\lambda \in \mathbb{R}^+$ is a margin between positive and negative pairs. This loss defines the *triplet constraint* as the requirement that the distance of negative pairs should be larger than the distance of positive pairs.

**Inference.** At inference time, WhosAI exploits an off-line step that consists in precomputing the *centroids* in $\mathcal{D}$ for each category $c_k \in \mathcal{C}$, defined as $\mathbf{c}_k = (1/|\mathcal{D}_k|) \sum_{D_i \in \mathcal{D}_k} \mathbf{h}_i$, where $\mathcal{D}_k$ denotes the subset of $\mathcal{D}$ containing data objects of category $c_k$.

Given a previously unseen data object $D$, WhosAI computes its embedding $\mathbf{h}$ (Eq. 1), which is then compared to each of the centroids in such a way that $D$ is assigned to the category $c_{k^*}$ that corresponds to the least distant centroid:

$$k^* = \arg\min_{k=1..M} d(\mathbf{h}, \mathbf{c}_k). \quad (3)$$

### 4.1 Optimizations

We discuss here a set optimization techniques as enhancements of key components in WhosAI, namely improved triplet mining, dynamic margin scheduling, and data corruption.

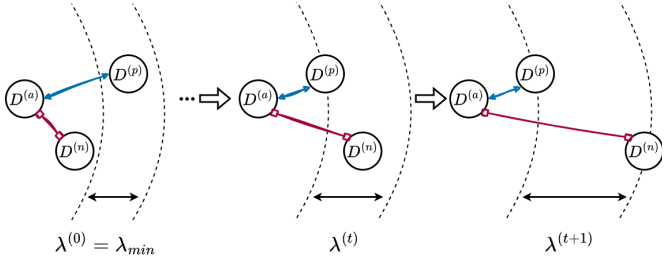

**Figure 2.** On the left, an example of violation at early training-stage of the triplet constraint, as the distance of the negative pair is not larger than the distance of the positive pair. As the training progresses (mid and right), the margin between positive and negative pairs dynamically increases, thus strengthening the fulfillment of the triplet constraint.

**Improving Triplet Mining.** A straightforward implementation of the triplet mining process involves gathering triplets before each training epoch and feeding batches of these triplets into the Triplet Network, essentially as an "offline" process. However, this approach might have two main drawbacks: (i) not all generated triplets may contain the valuable information needed for minimizing the loss (Eq. 2), and (ii) triplets regarded as "informative" in an earlier stage of training might quickly become "uninformative" as the model's weights undergo updates.

Within this view, it becomes crucial for the triplet mining process to prioritize an *online* identification of the most informative triplets for each training epoch. These should be the most unexpected ones, i.e., triplets that most violate the margin constraints enforced by the loss function. This strategy can improve the mining process as it enhances the generalization capabilities and training stability, and it makes training more efficient by avoiding the inclusion of the uninformative triplets.

The above requirements can effectively be fulfilled by the pair mining scheme adopted in the *multi-similarity miner* method [48]. Essentially, the pair mining consists in sampling informative pairs through the relative similarity between the negative and positive pairs sharing a common anchor. More specifically, a negative pair is selected as one having lower distance than the hardest positive pair (i.e., the one with the highest distance):

$$d(\mathbf{h}^{(a)}, \mathbf{h}^{(n)}) < \max_{D^{(p)}} d(\mathbf{h}^{(a)}, \mathbf{h}^{(p)}) + \varepsilon. \quad (4)$$

A positive pair is selected as one having higher distance than the hardest negative pair (i.e., the one with the lowest distance):

$$d(\mathbf{h}^{(a)}, \mathbf{h}^{(p)}) > \min_{D^{(n)}} d(\mathbf{h}^{(a)}, \mathbf{h}^{(n)}) - \varepsilon. \quad (5)$$

**Dynamic Margin Scheduling.** Another improvement we consider is to make the training of WhosAI progressively harder. Specifically, by *dynamically increasing* the margin $\lambda$ in our loss function (Eq. 2), we require the model to focus on harder negative pairs as the training goes on, in order to produce an enhanced separation between classes.

To this aim, inspired by curriculum-based learning [17], we revise the loss function with a dynamic margin that follows a linear schedule dependent on the training step time $t \geq 0$, which is defined as:

$$\lambda^{(t)} = \lambda_{\min} + \lambda_\Delta (t \bmod \delta), \quad (6)$$

where $\lambda_{\min} \in \mathbb{R}^+$ is the initial margin, $\lambda_\Delta \in \mathbb{R}^+$ denotes the margin increment, and $\delta$ represents the step size of the increment. The rationale of this formula is as follows. We begin with an initial, relatively low margin, $\lambda_{\min}$, to facilitate manageable gradients during early optimization; in fact, at early stage, a model can exhibit some discriminative ability, however, large margins during this stage would lead to excessively large gradients, hindering learning. As the optimization progresses and the distance constraints are enforced, the importance of loss-based gradients gradually diminishes. To prevent stagnation, the margin is hence periodically increased by $\lambda_\Delta$ every $\delta$ training steps. A visual representation of the process is shown in Figure 2.

Based on Eq. 6, our loss function becomes dynamic by integrating *dynamic margin scheduling* with the triplet loss:

$$\mathcal{L}^{(t)} = \sum_{\langle D^{(a)}, D^{(p)}, D^{(n)} \rangle} \max(d(\mathbf{h}^{(a)}, \mathbf{h}^{(p)}) - d(\mathbf{h}^{(a)}, \mathbf{h}^{(n)}) + \lambda^{(t)}, 0). \quad (7)$$

**Data Corruption.** In the context of language modeling, different strategies of data augmentation can be used in order to generate new training examples by perturbing existing input sequences. The general effect is to increase the diversity and variability of the training data, thereby improving the model's ability to generalize and robustness to different input variations. In this regard, we focus on the process of removing individual tokens, or groups of tokens, from a given input sequence and observing the impact on the model's output. Specifically, we consider the following operations, which have previously shown to be effective in improving the performance of PLMs in several tasks (e.g., [19, 27]):

- *token deletion*: given the token sequence $T_i = [\tau_{i,1}, \ldots, \tau_{i,|T_i|}]$, a token $\tau_{i,j}$ is removed with probability $p \sim U[0, 1]$;
- *span cropping*: given the token sequence $T_i = [\tau_{i,1}, \ldots, \tau_{i,|T_i|}]$, a token $\tau_{i,j}$ is selected as a starting index with probability $p_s \sim U[0, 1]$; then, for each sampled starting index, a span size $sz \sim U[0, |T_i| \times p_{span}]$ is also sampled, where $p_{span}$ indicates the relative size of the span w.r.t. the overall sequence size. Finally, the sampled spans of tokens are deleted from the sequence.

It is important to note that these operations aim to remove specific (sequences of) tokens from our input text instead of masking them. Accordingly, our PLMs are not required to reconstruct the missing tokens, as in masking-based tasks.

## 5 Experimental Methodology

### 5.1 Data

We used the publicly available benchmark dataset *TuringBench* [42, 41], which contains 200K news articles, where 10K are human-written and the other ones are machine-generated news articles equally distributed over 19 different AI text-generation models. From the human-written articles, originally collected from sources like CNN and with typical length of 200-400 words, the titles were used to prompt the 19 AI text-generators to generate 10K articles each. Table 1 summarizes the main characteristics of the dataset, providing details for the various subsets of data associated with the human category and each of the AI text-generator categories. It should be noted that TuringBench comes with a pre-defined split into train, validation and test sets. We will follow this setting, so as to fully compare with previous and future evaluation studies on TuringBench.

### 5.2 Assessment Criteria and Model Settings

To validate the performance of WhosAI in detecting and attributing AI-generated text, we resort to standard statistics based on the confusion matrices derived from testing WhosAI predictions w.r.t. the ground-truth under the Turing Test task and w.r.t. the ground-truth under the Author Attribution task, respectively. These include the weighted average (i.e., averaging over the support-weighted mean per class) of *precision* ($P$), *recall* ($R$), and $F_1$-*score* ($F_1$).

**Table 1.** Main characteristics of the TuringBench. Table adapted from [42].

| Generation model (subset) | Ref. | Avg #Words per document | Avg. #Sentences per document | Params |
|---|---|---|---|---|
| Human | – | 232.7 | 15.0 | – |
| GPT-1 | [34] | 316.7 | 10.5 | 117M |
| GPT-2$_{small}$ | [35] | 118.6 | 4.0 | 124M |
| GPT-2$_{medium}$ | [35] | 120.9 | 4.2 | 355M |
| GPT-2$_{large}$ | [35] | 119.7 | 4.1 | 774M |
| GPT-2$_{xl}$ | [35] | 117.8 | 4.1 | 1.5B |
| GPT-2$_{PyTorch}$ | NA | 178.9 | 7.03 | 344M |
| GPT-3 | [6] | 129.5 | 5.0 | 175B |
| GROVER$_{base}$ | [54] | 299.2 | 9.4 | 124M |
| GROVER$_{large}$ | [54] | 286.3 | 8.7 | 355M |
| GROVER$_{mega}$ | [54] | 278.9 | 9.2 | 1.5B |
| CTRL | [22] | 398.1 | 20.0 | 1.6B |
| XLM | [26] | 387.8 | 4.2 | 550M |
| XLNET$_{base}$ | [52] | 226.1 | 11.6 | 110M |
| XLNET$_{large}$ | [52] | 415.8 | 4.3 | 340M |
| FAIR$_{wmt19}$ | [32] | 221.2 | 14.6 | 656M |
| FAIR$_{wmt20}$ | [11] | 100.6 | 5.1 | 749M |
| TRANSFORMER$_{XL}$ | [12] | 211.7 | 9.8 | 257M |
| PPLM$_{distil}$ | [13] | 156.9 | 10.7 | 82M |
| PPLM$_{gpt2}$ | [13] | 188.9 | 11.9 | 124M |

We also account for distance-based quantitative criteria that express how well the learned space aligns with the predefined categorization of the training texts, in terms of *compactness* within same-category groups of objects and *separation* between groups of objects of different categories. To this purpose, by denoting with $sim(\cdot, \cdot)$ the cosine similarity function, we calculate the average pairwise similarity of the embeddings of objects sharing the same category:

$$intra(\mathcal{D}) = \frac{1}{|\mathcal{D}_k|} \sum_{c_k \in \mathcal{C}} \sum_{D_i, D_j \in \mathcal{D}_k} sim(\mathbf{h}_i, \mathbf{h}_j), \qquad (8)$$

and the average pairwise similarity of the embeddings of objects belonging to two different categories:

$$inter(\mathcal{D}) = \frac{1}{|\mathcal{D}_h||\mathcal{D}_k|} \sum_{c_h, c_k \in \mathcal{C}} \sum_{D_i \in \mathcal{D}_h, D_j \in \mathcal{D}_k} sim(\mathbf{h}_i, \mathbf{h}_j). \qquad (9)$$

Following the most widely used approaches to sentence embedding [36], we used BERT [15] as our reference PLM, leveraging its publicly available `bert-base-uncased` implementation hosted on the HuggingFace platform.[3] This has a total of 110M parameters, distributed across 12 Transformer-encoder layers with 12 attention heads, a vocabulary of 32K tokens, maximum length of context set to 512 tokens, and embedding size $f$ set to 768. We used the AdamW optimizer with learning rate of 1.0E-5, $(\beta_1, \beta_2)=(0.9, 0.99)$, and a weight decay, useful for regularization, of 0.01. We set the batch size to 32 elements, and the number of steps for each training procedure to 30K. Furthermore, we employ a linear learning rate scheduler with a 3000-step warm-up. As concerns the dynamic margin scheduling, we linearly distribute margin updates during the training process, by setting $\lambda_{\min} = 0.1$, and $\delta = 750$, which implies $\lambda_\Delta = \frac{750}{30000} = 0.025$. For the data corruption operations, we set the probability $p = 0.05$ for the *token deletion* (TD) function, and both the probabilities $p_s$ and $p_{span}$ to 0.05 for the *span cropping* (SC) function. Throughout our work, we used the cosine distance defined as $1 - sim(\cdot, \cdot)$.

Our experiments were carried out on a double 56-core Intel(R) Xeon(R) Gold 6258R CPU, with 256GB RAM and two NVIDIA GeForce RTX3090s, OS Ubuntu Linux 22.04 LTS.

[3] https://huggingface.co/google-bert/bert-base-uncased

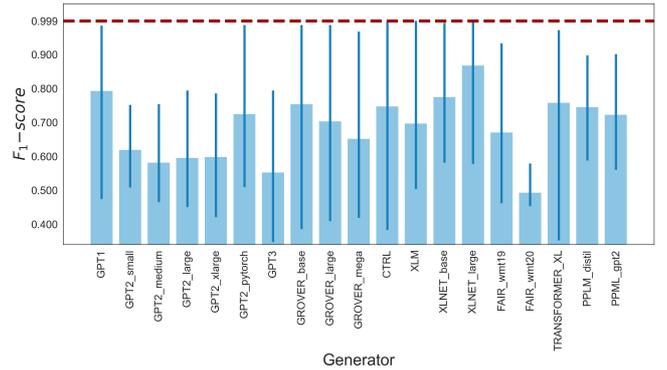

**Figure 3.** *Turing Test evaluation*: barchart of the average $F_1$ score from the TuringBench Leaderboard, for each TuringBench subset (i.e., generator). The horizontal red dashed line corresponds to the $F_1$ score achieved by WhosAI (best-performing setting) over the entire TuringBench test set.

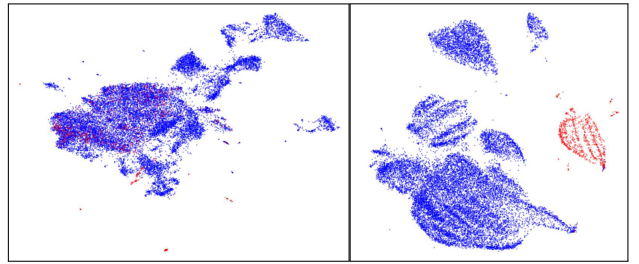

**Figure 4.** *Turing Test evaluation*: 2D UMAP visualization of the semantic space produced by WhosAI *before* (left) and *after* training (right). Colors denote human (red) vs. AI-generated (blue) texts.

## 6 Results

We organize our presentation of the results into three parts: the first two discuss quantitative and qualitative results on the Turing Test (TT) and Authorship Attribution (AA) tasks, respectively, achieved by WhosAI (best-performing setting) and competitors, whereas the third part focuses on evaluating the impact of our optimization strategies on the performance of WhosAI.

### 6.1 Turing Test

We start with evaluating WhosAI on the binary classification task, i.e., TT, aimed at recognizing whether a given piece of text originates from a human or any AI text-generator. As reported in Figure 3, the official TuringBench leaderboard[4] presents the $F_1$-scores for the TT under a *One-vs-One* approach, whereby one side of the comparison denotes "human" and the other one corresponds to each of the available AI-generators in TuringBench. It can be noticed that some generators are more easily detectable than others, resulting in substantial disparities in terms of average weighted $F_1$-scores.

By contrast, WhosAI is able to learn a deep semantic space for the whole set of generators at once. As a major result, WhosAI achieves an impressive $F_1$-score of 0.999 on the whole TT test set supplied by the TuringBench benchmark, setting a new best performance on the Turing Test. Our remarkable $F_1$-score is further corroborated by a qualitative analysis based on the visualization provided in Figure 4:[5] while at the beginning of the training the semantic representation directly induced by the PLM does not adequate separate the human and AI subspaces, the final trained WhosAI shows its ability to learn perfectly to recognize the two classes for the Turing Test. This couples with the remarkable results reported in Table 4, whereby the average pairwise similarity between embeddings of objects sharing the same category, resp. belonging to different categories, is of 0.931, resp. -0.808. These highlight a notable coherence within each category and a clear separation between categories.

[4] Available at https://turingbench.ist.psu.edu/
[5] Figures 4–5 were obtained by transforming the WhosAI embeddings using UMAP [30] default parameters with 2 components and cosine distance.

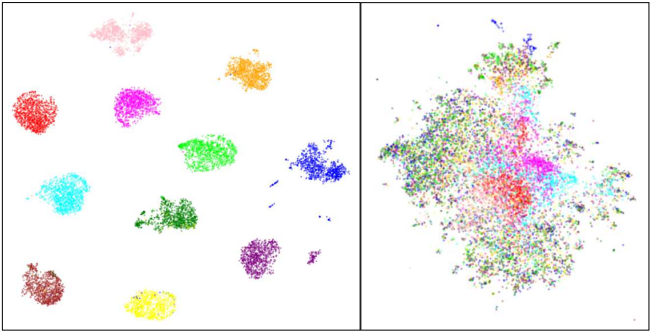

**Figure 5.** *Authorship Attribution evaluation*: 2D UMAP visualization of the semantic space produced by WhosAI (left) and SBERT (right). Colors denote human (blue) and the various AI text generators.

### 6.2 Authorship Attribution

We discuss our evaluation of WhosAI on the Author Attribution (AA) task, aimed at deciding the authorship of a text, being a human or one of the AI text-generators in TuringBench.

Our first remarkable finding derives from a comparison between WhosAI results against those reported on the TuringBench leaderboard for the AA task, whose top-5 best-performing models are shown in Table 2. RoBERTa [29] with a multi-class classification setting turns out to be the best model in the leaderboard for the AA task, with a $F_1$ score of 0.811, followed by other BERT-based approaches, as well as the official OpenAI detector and machine learning-based models. The winner method from the leaderboard is however outperformed by WhosAI, which achieves a striking average weighted $F_1$ score, precision and recall of 0.990, thus demonstrating almost perfect capabilities of authorship prediction.

Table 3 offers insights into the prediction performance of WhosAI w.r.t. each of the generator categories corresponding to the largest-size versions of the AI models. Results show extremely robustness of WhosAI, as it achieves $F_1$ score at least 0.960, and above 0.99 in 7 out of 10 cases. It should also be noticed that precision and recall are always comparable or very close to each other, thus indicating an equal capability of avoiding both types of statistical errors.

As previously found for the Turing Test task, the striking $F_1$ scores achieved by WhosAI couple with an evidence of highest cohesiveness and separation of the subspaces associated with the various text authorships, as visually shown in Fig. 5 (left); quantitatively, this corresponds to an average pairwise similarity between embeddings of objects sharing the same category, resp. belonging to different categories, of 0.938, resp. -0.012 (cf. Table 5).

It is worth noting that the outstanding performance by WhosAI in the AA task is not paired by a state-of-the-art sentence-embedding method for semantic-similarity-related tasks like SBERT [36], based on a Siamese network using BERT at its core: indeed, as shown in Fig. 5 (right), the intra-class cohesiveness and inter-class separation of the semantic space learned by SBERT are clearly worse than those achieved by WhosAI.

**Table 2.** *Authorship Attribution evaluation*: Results achieved by WhosAI vs. the top-5 models from the TuringBench Leaderboard (https://turingbench.ist.psu.edu/).

| Detection method | $P$ | $R$ | $F_1$ |
|---|---|---|---|
| WhosAI | **0.990** | **0.990** | **0.990** |
| RoBERTa | 0.821 | 0.813 | 0.811 |
| BERT | 0.803 | 0.802 | 0.800 |
| BERTAA | 0.780 | 0.775 | 0.776 |
| OpenAI detector | 0.781 | 0.781 | 0.774 |
| SVM (3-grams) | 0.712 | 0.722 | 0.715 |

**Table 3.** *Authorship Attribution evaluation*: Summary of per-category results achieved by WhosAI.

| Generation model (class) | $P$ | $R$ | $F_1$ | support |
|---|---|---|---|---|
| Human | 0.999 | 0.992 | 0.995 | 975 |
| GPT-1 | 1.000 | 1.000 | 1.000 | 993 |
| GPT-$2_{xl}$ | 0.950 | 0.970 | 0.960 | 993 |
| GPT-3 | 0.977 | 0.959 | 0.968 | 894 |
| GROVER$_{mega}$ | 0.999 | 0.997 | 0.998 | 894 |
| CRTL | 0.998 | 0.999 | 0.999 | 1000 |
| XLM | 1.000 | 1.000 | 1.000 | 973 |
| XLNET$_{large}$ | 0.999 | 1.000 | 1.000 | 1000 |
| FAIR$_{wmt20}$ | 0.983 | 0.986 | 0.984 | 993 |
| TRANSFORMER$_{XL}$ | 0.996 | 0.994 | 0.995 | 991 |
| PPLM$_{gpt2}$ | 0.990 | 0.992 | 0.991 | 975 |
| Overall | 0.990 | 0.989 | 0.989 | 10681 |

### 6.3 Sensitivity Analysis

**Turing Test.** Table 4 reports the results achieved by WhosAI on the TT task by varying the framework settings according to the various optimizations discussed in Section 4.1.

At first glance, we notice that WhosAI can solve the TT task almost perfectly – with $P$, $R$, $F_1$ always above 0.996 – regardless of specific optimizations. This remarkable finding, coupled with the visual evidence of the semantic space representation displayed in Figure 4, indicate that WhosAI excels in distinguishing between human authors and AI text-generators, even when equipped with the simplest configuration.

While the classification performance criteria have indeed only slight fluctuations by varying the framework settings, different behaviors of WhosAI appear to be more evident in terms of compactness ($intra$) and, especially, separation ($inter$), with the former consistently above 0.846 and the latter that can vary from 0.35 to -0.81. In particular, applying data corruption techniques can affect the distance-based criteria: in fact, by using either token deletion and span cropping, we notice a worse (i.e., higher) similarity between embeddings of objects pertaining to different categories, whereas the similarity between embeddings of objects from the same category remains coherent. This suggests that corrupting the data in input to the TT predictor might impact particularly on some of the tokens that are discriminative of the text-generators, being human or AI models.

More importantly for our TT evaluation, we assessed the effect on the WhosAI performance due to the presence of texts that were generated by the same AI architecture yet with different *parameter sizes* (cf. Table 1). To this aim, we focused on comparing the performance of WhosAI when keeping *all* instances generated by the same AI text-generation architecture, and only the subsets corresponding to either the largest model or the smallest model of that AI architecture available in TuringBench.

As reported in Table 4, the variation of the model subset mainly impacts on the separation between embeddings of objects pertaining to different classes: keeping all instances from differently sized mod-els from the same architecture can bring additional discriminative information helping WhosAI better separate the human-generated texts from the ones generated by all AI models. Conversely, maintaining only either the largest-model or smallest-model subsets might lead to slightly improved intra-class cohesiveness, hinting at a reduced noise affecting characterizing tokens.

**Table 4.** *Turing Test evaluation*: Results by varying the setting of WhosAI. Most preferable setting is bolded.

| Triplet Mining | Dynamic Margin | Data Corrupt. | Generator subset | $P$ | $R$ | $F_1$ | $inter$ | $intra$ |
|---|---|---|---|---|---|---|---|---|
| ✗ | ✗ | ✗ | **All** | **0.999** | **0.999** | **0.999** | **-0.808** | **0.931** |
| ✓ | ✗ | ✗ | All | 0.999 | 0.999 | 0.999 | -0.805 | 0.914 |
| ✓ | ✓ | ✗ | All | 0.999 | 0.999 | 0.999 | 0.275 | 0.941 |
| ✓ | ✓ | ✗ | Largest | 0.999 | 0.999 | 0.999 | 0.050 | 0.964 |
| ✓ | ✓ | SC | Largest | 0.998 | 0.998 | 0.998 | 0.275 | 0.892 |
| ✓ | ✓ | TD | Largest | 0.999 | 0.999 | 0.999 | 0.287 | 0.955 |
| ✓ | ✓ | ✗ | Smallest | 0.998 | 0.998 | 0.998 | 0.147 | 0.951 |
| ✓ | ✓ | SC | Smallest | 0.996 | 0.996 | 0.996 | 0.353 | 0.846 |
| ✓ | ✓ | TD | Smallest | 0.998 | 0.998 | 0.998 | 0.321 | 0.958 |

**Table 5.** *Authorship Attribution evaluation*: Results by varying the setting of WhosAI. Most preferable setting is bolded.

| Triplet Mining | Dynamic Margin | Data Corrupt. | Generator subset | $P$ | $R$ | $F_1$ | $inter$ | $intra$ |
|---|---|---|---|---|---|---|---|---|
| ✗ | ✗ | ✗ | All | 0.769 | 0.775 | 0.763 | -0.030 | 0.921 |
| ✗ | ✓ | ✗ | All | 0.767 | 0.774 | 0.761 | -0.031 | 0.923 |
| ✓ | ✗ | ✗ | All | 0.782 | 0.789 | 0.779 | 0.208 | 0.891 |
| ✓ | ✓ | ✗ | **Largest** | **0.990** | **0.990** | **0.990** | **-0.012** | **0.938** |
| ✓ | ✓ | SC | Largest | 0.985 | 0.985 | 0.985 | 0.048 | 0.938 |
| ✓ | ✓ | TD | Largest | 0.989 | 0.989 | 0.989 | 0.028 | 0.938 |
| ✓ | ✓ | ✗ | Smallest | 0.989 | 0.989 | 0.989 | 0.003 | 0.933 |
| ✓ | ✓ | SC | Smallest | 0.988 | 0.988 | 0.988 | 0.069 | 0.939 |
| ✓ | ✓ | TD | Smallest | 0.991 | 0.991 | 0.991 | 0.038 | 0.936 |

**Authorship Attribution.** Analogously to the previous analysis, Table 5 summarizes the results achieved by WhosAI on the AA task based on different settings. We notice that WhosAI obtains an $F_1$ score above 0.980, also with remarkable distance-based scores, in 6 out of 9 configurations, which correspond to testing on texts from a particular model-size variant of an AI text generator, rather than testing on all texts from the different variants of a model.

This prompted us to investigate the similarity between the centroids of the subsets corresponding to the various categories (i.e., model instances), as reported in Fig. 6 (left). Notably, looking at the diagonal blocks in the heatmap, WhosAI consistently learns identical centroids (i.e., cosine similarity equal to 1) for the different instances of the same AI generation architecture, regardless of the parameter size. While this suggests that a single instance of a given architecture may be enough in a contrastive setting to characterize a whole family of AI generators, it introduces lots of noise when distinguishing between two or more instances of the same architecture.

If we focus on the largest-model variants of the AI generators, which should make our AA task more challenging as it is supposed that more parameters enable the model to capture more knowledge providing it with better generation capabilities, we still find in Fig. 6 (right) low values of inter-class similarity. Analogous results (not shown) are achieved for the smallest model instances, which may be preferable in resource-constrained scenarios.

Furthermore, considering the impact of the different optimizations, the triplet mining and the dynamic margin scheduling lead to performance improvements, while the data corruption methods appear to worsen the separation of the learned embedding subspaces by affecting tokens crucial for discriminating text-generators.

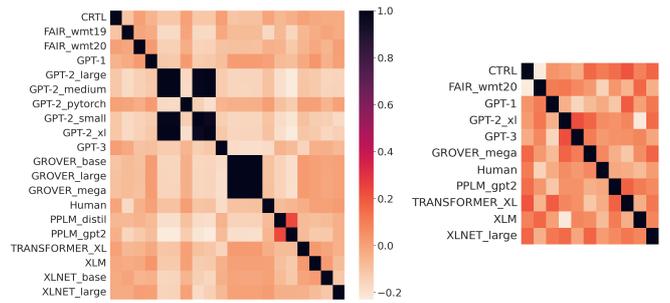

**Figure 6.** Cosine similarity between centroids of the different categories for all (left), resp. the largest (right), generators. Darker colors indicate higher similarity.

## 7 Conclusions and Future Work

We tackled the challenge of detecting and attributing AI-generated text through WhosAI, a novel PLM-based framework that leverages contrastive learning to induce a semantic similarity space texts written by humans or AI text-generation models. This similarity space is efficiently exploited at inference time by means of a nearest centroid classifier to predict the authorship of unlabeled texts. Extensive experimentation on the well-known *TuringBench* dataset has revealed state-of-the-art performances of WhosAI on both TT and AA tasks. Furthermore, WhosAI comes with several key advantages: (i) it can be applied straightforwardly without altering texts or accessing models' internals, (ii) it can be adapted to a number of AI text-generators without needing model-specific adjustments, and (iii) it is model-agnostic and scalable for easy integration of novel AI text-generators. Remarkably, such empirical evidence of outstanding performance of WhosAI holds despite our choice of PLM in the experimental evaluation refers to the baseline BERT model.

There are important directions to explore. Particularly, we will evaluate WhosAI on other types of written texts than those available in TuringBench. We aim to compare WhosAI with advanced yet commercially licensed AI detection tools (e.g., GPTZero). Also, we will investigate explainability aspects of WhosAI in order to unveil which features are determinant to characterize and which to discriminate text originators.

**Remarks on the carbon footprint of WhosAI.** As a supplementary analysis, we investigated the environmental impact of WhosAI. Based on our selected reference PLM, which has 109.48M parameters, we estimate an inference cost of 290.17 GFLOPS for a single-element batch with a sequence length of 512, and a training cost of 835.8 PFLOPS assuming 32 batch size, 512 sequence length, and 30K training steps. With an average training time for all WhosAI configurations of ∼8 hours on a 350W Nvidia GeForce RTX 3090 GPU, we estimate an energy consumption of 2.8 kWh per training run. With an average carbon efficiency factor of 0.432 kg/kWh, a single training run is associated with 1.21 kg of $CO_2$ emissions, which extends to 50.4 kWh of consumed energy and 21.78 kg of $CO_2$ equivalent emissions for running all of our experiments.[6]

### Acknowledgements
AT, resp. LLC, was supported by project "Future Artificial Intelligence Research (FAIR)" spoke 9 (H23C22000860006), resp. project SERICS (PE00000014), both under the MUR National Recovery and Resilience Plan funded by the EU - NextGenerationEU.

---

[6] For this analysis, we used `calflops` (https://github.com/MrYxJ/calculate-flops.pytorch) and `impact` (https://github.com/mlco2/impact).